# Catch and Prolong: recurrent neural network for seeking track-candidates

*Dmitriy* Baranov[1], *Gennady* Ososkov[1], *Pavel* Goncharov[2,*] and *Andrei* Tsytrinov[2]

[1]Joint Institute for Nuclear Research, 6 Joliot-Curie street, Dubna, Moscow region, Russia
[2]Sukhoi State Technical University of Gomel, October Ave. 48, Gomel, 246746 Republic of Belarus

**Abstract.** One of the most important problems of data processing in high energy and nuclear physics is the event reconstruction. Its main part is the track reconstruction procedure which consists in looking for all tracks that elementary particles leave when they pass through a detector among a huge number of points, so-called hits, produced when flying particles fire detector coordinate planes. Unfortunately, the tracking is seriously impeded by the famous shortcoming of multiwired, strip and GEM detectors due to appearance in them a lot of fake hits caused by extra spurious crossings of fired strips. Since the number of those fakes is several orders of magnitude greater than for true hits, one faces with the quite serious difficulty to unravel possible track-candidates via true hits ignoring fakes. We introduce a renewed method that is a significant improvement of our previous two-stage approach based on hit preprocessing using directed K-d tree search followed a deep neural classifier. We combine these two stages in one by applying recurrent neural network that simultaneously determines whether a set of points belongs to a true track or not and predicts where to look for the next point of track on the next coordinate plane of the detector. We show that proposed deep network is more accurate, faster and does not require any special preprocessing stage. Preliminary results of our approach for simulated events of the BM@N GEM detector are presented.

Keywords: tracking, BM@N experiment, GEM detectors, directed search, recurrent neural networks

## 1 Introduction

Event reconstruction in particles track detectors is the very important problem in modern high energy and nuclear physics (HENP). One of its significant parts is the tracking procedure that consists in finding tracks among a great number of so-called hits produced by flying particle interaction with sequential coordinate planes of tracking detectors. This procedure is especially difficult for modern HENP experiments with heavy ions where detectors register events with very high multiplicity.

---

[*]e-mail: kaliostrogoblin3@gmail.com

At the same time, while working on BM@N experiment [1] we have faced with the famous shortcoming of GEM strip detector, when a great amount of fake hits appears besides of real hits because of extra spurious crossings of strips. The number of those fakes is greater for some order of magnitude than for true hits (see [2], for example).

A common method for dealing with track reconstruction is the combinatorial Kalman filter, which have been used with great success in HENP experiments for years [3]. However, the initialization of Kalman filter is cumbersome process, because of a really vast search of hits needed to obtain so-called "seeds", i.e. initial approximations of track parameters of charged particles.

Taking into account sequential nature of any tracking detector and the consideration that machine learning algorithms could make a great contribution to the tracking problem due to their capability to model complex non-linear data dependencies, we have proposed a two-step approach to the particle track reconstruction based on deep learning methods [2].

In the given work we will improve this tracking model in order to overcome its disadvantages by combining two-steps in one end-to-end trainable deep learning network. We will show, that the new approach outperforms its progenitor in all aspects: accuracy, processing time, – and it doesn't need a preprocessing.

## 2 Previous study

In [2] we proposed a directed search of track-candidates on the first step of the track reconstruction algorithm. It can be simply described as a procedure that goes through reconstructed hit points at every station, starting from the first one and extending current track candidates by one hit at every station also taking into account possible target coordinates.

To speed up such a preprocessing procedure, we took into account the direction of magnetic field to arrange the search in two coordinate projections simultaneously: in one tracks are represented as straight lines and in another as circles. K-d tree data structure is used to bind the searching area of each continuation of track-candidates.

On YoZ projection, tracks are almost straight lines. Thus, merely by sorting hit indices array by y coordinate, making a confidence interval and executing a binary search we can exclude all hits that are not in the confidence interval. For the XoZ projection, where the tracks look like circles, we enable limited rotational component. The limitation is on change in rotation – it should not change substantially.

After doing the first step of our tracking we obtain a bunch of track-candidates, which should be divided in two groups: real tracks and, so named, ghost tracks formed by fakes and, possibly, by parts of different tracks. To addressing the classification task, we utilized a deep recurrent neural network (RNN) as a classifier.

RNN was chosen due to its dynamical structure, considering the sequential nature of the tracking problem. Hits belonging to some of tracks are indicated by three coordinates – features and situated on sequential stations along a particle passage through the detector. Thus, a track-candidate presents as a sequence of hits. RNN can process sequences of different length, which is useful for processing tracks of particles with low energy.

After an extensive study of neural model selection we found that the best results of validation efficiency can be obtained by the combination of one convolutional layer and two GRU [4] layers one after another. The full algorithm description one can find in [2].

Testing efficiency equals to 97.5%. Trained RNN can process 6500 track-candidates in one second on the single Nvidia Tesla M60.

## 3 Killing two birds with one RNN

In spite of all, our two-step solution doesn't have a suitable performance due to spending a lot of time while building K-d tree structure for every event always from scratch. Thus, we decided to come up with a model, that can predict the probability whether or not seeds belong to true track (looks like trainable version of admissibility function in [2]) and the searching area for the continuation of track-candidate on the next coordinate plane based on input set of points.

### 3.1 Model architecture

To build a new model we changed considerably our deep RNN classifier from the previous two-stage approach. These changes required the following: to replace the bidirectional GRU layer with the regular one-directional, to reduce the number of neurons by half in every recurrent layer, to remove the dropout layers and, eventually, to add to the output layer a special regression part needed to predict on the next coordinate plane an elliptical area, where to search for the track-candidate continuation. This regression part consists of four neurons, two of which with linear activation functions determine the ellipse center and another two – define the semiaxes of that ellipse. Neurons, which predict the semiaxes of ellipses have softplus activations [5] defined as follows:

$$f(x) = log(1 + e^x) \qquad (1)$$

Our new model, presented on the figure 1, takes, as input, sets with different number of hits in them: from two (target and zero station's hit) up to N, where N is the number of detector readout stations. For sequences with two hits even one cannot deduce the probability whether this sequence is a track or not, because two points indicate only a direction, where we should find the continuation of this set, thus the output will be presented by the regression part only. Whilst for the sequences of the maximum length N the output consists of the only one sigmoid neuron, since we do not need to seek for the next point on the very last station.

### 3.2 Cost function

To train the proposed new model we introduce an ad hoc cost function, defined as follows:

$$J = max(\lambda_1, 1 - p)\, FL(p, p') + p\left(\lambda_2 \sqrt{\left(\frac{x-x'}{R1}\right)^2 + \left(\frac{y-y'}{R2}\right)^2} + \lambda_3 R1R2\right), \qquad (2)$$

where $\lambda_{1\text{-}3}$ are the weights for each part of equation; $p$ is the label that indicates whether or not the set of input points belongs to true track; $p'$ is the probability of track/ghost was predicted by network; $x'$, $y'$ are coordinates of the center of ellipse, predicted by network; $x$, $y$ are coordinates of the next point of the true track segment; $R1$, $R2$ are semiaxes of the ellipse; $FL(p, p')$ is a balanced focal loss [6] with a weighting factor $\alpha \in [0, 1]$. Introduction of such weighting factor is the common method for addressing any class imbalance. We set $\alpha = 0.95$. The idea of focal loss function is to down-weight easy examples and thus focus training on hard negatives. The focusing parameter $\gamma$ (we set it to 2) smoothly adjusts the rate at which easy examples are down-weighted.

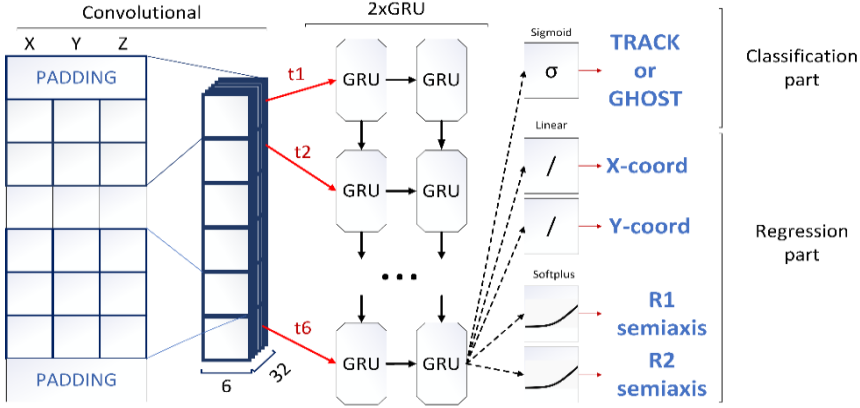

**Fig. 1.** The main scheme of the deep RNN for tracking. Worth to note that the size of output depends on input's size, e.g. for sequences of size two the output contains only the regression part and for maximum size contains the only sigmoid neuron.

While dealing with fake tracks we do not need to consider the error attended to ellipse size and position, for that reason we put the label $p$ before the second part of equation. Thus when the label is 0 (= ghost), we multiply the error of the regression part by zero. And, on the contrary, the $max(\lambda_1, 1 - p)$ is utilized to focus on the classification error, when the regression part is turned off.

Summing up briefly all above mentioned, the equation (2) consists of three parts corresponding to $\lambda_{1-3}$: classification loss, point in ellipse loss, ellipse size penalization.

### 3.3 Dataset and training setup

To prepare the dataset for training and evaluating the new model, we simulated 15 thousands of events with 20-30 tracks per event using Box generator [7]. Thereafter, we ran K-d tree search for obtaining the bunch of labelled track-candidates (seeds) based on knowledge which of them are true tracks and which are not. Eventually, we obtained 82 677 real tracks and 695 887 ghosts.

During every training iteration, the training set of seeds is splitted in three groups of track-segments containing different number of points (from 2 to 6). For each of these seeds the network should predict the probability whether that set of points belongs to a true track (except 2 points) and also an elliptic area, where to search for a track continuation (except 6 points).

Deep RNN have been trained with $\lambda_1 = 0.5$, $\lambda_2 = 0.35$, $\lambda_3 = 0.15$, $\alpha = 0.95$, $\gamma = 2$ for 50 epochs with batch size equals to 128 and Adam optimization method. The loss value on the test subset of dataset after training stabilized on 0.019.

Worth to note, that each of track-candidates in dataset was labelled by K-d tree as potential track, so one can see that the sinus criterion in [2] is too rough.

## 4 Results

Results of the model evaluation on the test subset of data (250K of seeds with factor of 1:10 – one true track opposite to ten ghosts) are presented in table 1. We have leveraged several metrics to test the model's correctness for predictions of inputs' belongings to true tracks. They are known in statistics as accuracy, precision and recall [9].

**Table 1.** The results of applying several metrics on the trained network for different input length

|  | 3 points | 4 points | 5 points |
|---|---|---|---|
| **Recall** | 98.2% | 99.0% | 98.3% |
| **Precision** | 49.0% | 57.0% | 70.0% |
| **Accuracy** | 88.0% | 92.0% | 95.2% |
| **Ellipse square** | 1.67 cm$^2$ | 1.64 cm$^2$ | 1.91 cm$^2$ |

Accuracy (also known as efficiency) is the fraction of predictions our model got right, but it becomes useless while dealing with imbalanced dataset. Therefore, precision and recall metrics are used in these cases as more informative. Precision tells us how many of the objects classified as true tracks were correct. Recall means how many of the objects that should be marked as true tracks were actually selected. Thus, recall expresses the ability to find all true tracks in a dataset, while precision expresses the proportion of data, our model says was true, actually were true tracks.

In addition to classification metrics, we have supplemented the table with the row of ellipse squares sizes. Trained RNN can vary the size of predicted area for searching depending on the length of input sequence and angle of circle rotation. In the hottest region of the station 0 (the smallest station with the area of size 64x41 cm$^2$) the average number of hits, located on the area the size of predicted ellipse, is 1.65 hits (for 100k events).

In our study we speed up significantly our deep network calculations by using multicore computational systems and multiprocessor graphics cards or GPUs via the facilities provided by the JINR supercomputer GOVORUN [8]. The speed of the test run has reached 3 483 608 track-candidate/sec on 2x Tesla V100.